\documentclass{article}
\usepackage{eaclap}
\newcommand{\etal}{{\it et al.}}

\newcommand{\parseval}{{\sc parseval}}
\newcommand{\susanne}{{\sc susanne}}
\newcommand{\ptb}{{\sc ptb}}
\title{\vspace{-0.5in}\noindent
{\normalsize\it In Proceedings of the EACL workshop on Linguistically Interpreted Corpora (LINC), Bergen, Norway, June 1999
}\\
\vspace{6mm}
{\Large Corpus Annotation for Parser Evaluation}}

\author{John Carroll, {\bf Guido Minnen}\\
        Cognitive and Computing Sciences \\ University of Sussex\\
        Brighton BN1 9QH, UK\\
 \{johnca,guidomi\}@cogs.susx.ac.uk\\
\And
        Ted Briscoe\\
 Computer Laboratory\\
 University of Cambridge\\
 Pembroke Street, Cambridge CB2 3QG, UK\\
 ejb@cl.cam.ac.uk}

\begin{document}
\maketitle
\vspace{-0.5in}
\begin{abstract}

We describe a recently developed corpus annotation scheme for
evaluating parsers that avoids shortcomings of current methods. The scheme
encodes grammatical relations between heads and dependents, and has been used
to mark up a new public-domain corpus of naturally occurring English text. We
show how the corpus can be used to evaluate the accuracy of a robust parser,
and relate the corpus to extant resources.

\end{abstract}

\section{Introduction}

The evaluation of individual language-processing components forming
part of larger-scale natural language processing (NLP) application systems has
recently emerged as an important area of research (see e.g.\ Rubio, 1998;
Gaiz\-auskas, 1998). A syntactic parser is often a component
of an NLP system; a reliable technique for comparing and assessing the relative
strengths and weaknesses of different parsers (or indeed of different versions
of the same parser during development) is therefore a necessity.

Current methods for evaluating the accuracy of syntactic parsers are based on
measuring the degree to which parser output replicates the analyses assigned to
sentences in a manually annotated test corpus. Exact match between the parser
output and the corpus is typically not required in order to allow different
parsers utilising different grammatical frameworks to be compared. These
methods are fully {\it objective} since the standards to be met and
criteria for testing whether they have been met are set in advance.

The evaluation technique that is currently the most widely-used was proposed by
the Grammar Evaluation Interest Group (Harrison \etal,
1991; see also Grishman, Macleod \& Sterling, 1992), and is often known as
`\parseval'. The method compares phrase-structure bracketings produced by the
parser with bracketings in the annotated corpus, or
`treebank'\footnote{Subsequent evaluations using \parseval\ (e.g.\ Collins,
1996) have adapted it to incorporate constituent labelling information as well
as just bracketing.} and computes the number of bracketing matches $M$ with
respect to the number of bracketings
$P$ returned by the parser (expressed as precision $M/P$) and with respect to
the number $C$ in the corpus (expressed as recall $M/C$), and the mean number
of `crossing' brackets per sentence where a bracketed sequence from the parser
overlaps with one from the treebank and neither is properly contained in the
other.

Advantages of \parseval\ are that a relatively undetailed (only bracketed),
treebank annotation is required, some level of cross
framework/system comparison is achieved, and the measure is moderately
fine-grained and robust to annotation errors. However, a number of
disadvantages of \parseval\ have been documented recently. In particular,
Carpenter \& Manning (1997) observe that sentences in the Penn Treebank
(\ptb; Marcus, Santorini \& Marc\-in\-kie\-wicz, 1993) contain relatively few
brackets, so analyses are quite `flat'. (The same goes for the other
treebank of English in general use, \susanne; Sampson, 1995). Thus
crossing bracket scores are likely to be small, however good or bad the
parser is. Carpenter \& Manning also point out that with the adjunction
structure the \ptb\ gives to post noun-head modifiers {\it (NP (NP the man) (PP
with (NP a telescope)))}, there are zero crossings in cases where the VP
attachment is incorrectly returned, and {\it vice-versa}. Conversely,
Lin (1995) demonstrates that the crossing brackets measure can in some cases
penalise mis-attachments more than once; Lin (1996) argues that a high score
for phrase boundary correctness does not guarantee that a reasonable semantic
reading can be produced. Conversely, many phrase boundary disagreements stem
from systematic differences between parsers/ grammars and corpus annotation
schemes that are well-justif\-ied within the context of their own theories.
\parseval\ does attempt to circumvent this problem by the removal from
consideration of bracketing information in constructions for which agreement
between analysis schemes in practice is low: i.e.\ negation, auxiliaries,
punctuation, traces, and the use of unary branching structures.

However, in general there are still major problems with compatibility between
the annotations in treebanks and analyses returned by parsing systems using
manually-developed generative grammars (as opposed to grammars acquired
directly from the treebanks themselves). The treebanks have been construct\-ed
with reference to sets of informal guidelines indicating the type of
structures to be assigned. In the absence of a formal grammar controlling
or verifying the manual annotations, the number of different structural
configurations tends to grow without check. For example, the \ptb\ implicitly
contains more than 10000 distinct context-free productions, the majority
occurring only once (Charniak, 1996). This makes it very difficult to
accurately map the structures assigned by an independently-developed
grammar/parser onto the structures that appear (or should appear) in the
treebank. A further problem is that the \parseval\ bracket precision measure
penalises parsers that return more structure than the treebank annotation,
even if it is correct (Srinivas, Doran \& Kulick, 1995). To be able to use
the treebank and report meaningful \parseval\ precision scores such parsers
must necessarily `dumb down' their output and attempt to map it onto (exactly)
the distinctions made in the treebank\footnote{Gaizauskas, Hepple \& Huyck
(1998) propose an alternative to the \parseval\ precision measure to address
this specific shortcoming.}. This mapping is also very difficult to specify
accurately. \parseval\ evaluation is thus {\it objective}, but the results
are not {\it reliable}.

In addition, since \parseval\ is based on measuring similarity between
phrase-structure trees, it cannot be applied to grammars
which produce dependency-style analyses, or to `lexical' parsing frameworks
such as finite-state constraint parsers which assign syntactic functional
labels to words rather than producing hierarchical structure.

To overcome
the \parseval\ grammar/treebank mismatch problems outlined above, Lin (1995)
proposes evaluation based on dependency structure, in which phrase structure
analyses from parser and treebank are both automatically converted into sets
of dependency relationships. Each such relationship consists of a modifier, a
modifiee, and optionally a label which gives the type of the
relationship. Atwell (1996),
though, argues that transforming standard constitu\-ency-bas\-ed analyses
into a dependency-based representation would lose certain kinds of
grammatical information that might be important for subsequent processing,
such as `logical' information (e.g.\ location of traces, or moved
constituents). Srinivas, Doran, Hockey \& Joshi (1996) describe a related
technique which could also be applied to partial (incomplete) parses, in
which hierarchical phrasal constituents are flattened into chunks and the
relationships between them are indicated by dependency links. Recall and
precision are defined over dependency links.

The TSNLP (Lehmann \etal, 1996) project test suites (in English, French and
German) contain depend\-ency-based annotations for some sentences; this
allows for ``generalizations over potentially controversial phrase structure
configurations'' and also mapping onto a specific
constituent structure. No specific annotation standards or evaluation measures
are proposed, though.

\section{Grammatical Relation Annotation}\label{sec-scheme}

In the previous section we argued that constituency-based evaluation
for parser evaluation has serious short\-comings\footnote{Note that the issue
we are concerned with here is parser {\it evaluation}, and we are not making
any more general claims about the utility of con\-stituency-based treebanks
for other tasks, such as statistical parser
training or in quantitative linguistics.}. In this section we outline a
recently-proposed annotation scheme based on a dependency-style analysis, and
compare it to other related schemes. In the next section we describe a
10K-word test corpus that uses this scheme, and also how it may be used to
evaluate a robust parser.

Carroll, Briscoe \& Sanfilippo (1998) describe an annotation scheme in which
each sentence in the corpus is marked up with a set of grammatical relations
(GRs), specifying the syntactic dependency which holds between each head and
its dependent(s). The annotation scheme is application-independent, and takes
into account language phenomena in English, Italian, French and German. The
scheme is based on EAGLES lexicon/syntax working group standards (Sanfilippo
\etal, 1996), but refined within the EU 4th Framework SPARKLE project (see
$<$http://www.ilc.pi.cnr.it/sparkle/wp1-prefinal$>$) extending the set of
relations proposed there.

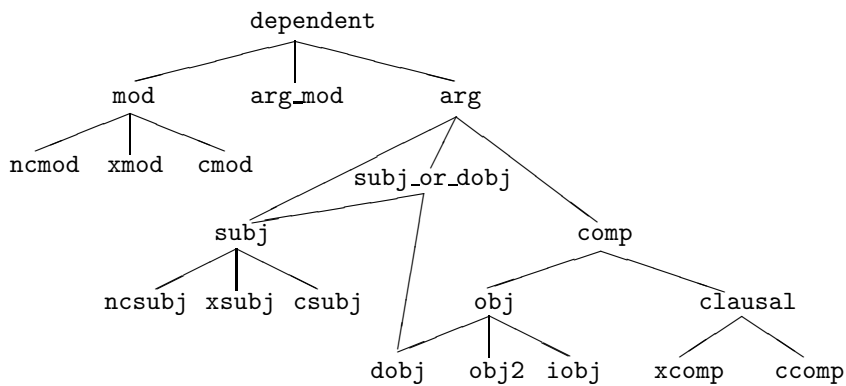
\begin{figure*}
\centering
{\tt    \setlength{\unitlength}{0.65pt}
\begin{picture}(480,230)
\thinlines    \put(354,88){\line(-3,-1){66}}
              \put(354,88){\line(3,-1){72}}

              \put(436,50){\line(3,-2){36}}
              \put(436,50){\line(-3,-2){36}}

              \put(289,50){\line(2,-1){47}}
              \put(289,50){\line(0,-1){24}}
              \put(289,50){\line(-5,-2){53}}

              \put(142,89){\line(2,-1){47}}
              \put(142,89){\line(0,-1){24}}
              \put(142,89){\line(-2,-1){47}}

              \put(270,166){\line(4,-3){82}}
              \put(270,166){\line(-1,-2){15}}
              \put(270,166){\line(-2,-1){120}}

              \put(251,121){\line(-1,-6){15}}
              \put(251,121){\line(-6,-1){100}}

              \put(80,168){\line(5,-2){55}}
              \put(80,168){\line(0,-1){24}}
              \put(80,168){\line(-5,-2){55}}

              \put(176,210){\line(4,-1){93}}
              \put(176,210){\line(0,-1){24}}
              \put(176,210){\line(-4,-1){93}}

              \put(150,218){dependent} 

              \put(70,174){mod}
              \put(150,174){arg\_mod}
              \put(260,174){arg} 

              \put(210,126){subj\_or\_dobj} 

              \put(10,134){ncmod}
              \put(67,134){xmod}
              \put(119,134){cmod}

              \put(129,94){subj}
              \put(340,94){comp} 

              \put(65,54){ncsubj}
              \put(124,54){xsubj}
              \put(175,54){csubj}

              \put(280,54){obj} 
              \put(411,54){clausal}

              \put(220,14){dobj}
              \put(277,14){obj2}
              \put(322,14){iobj}

              \put(385,14){xcomp}
              \put(455,14){ccomp}
\end{picture}}
\caption{The GR hierarchy.
}
\label{grs-used}
\end{figure*}

For brevity, we give an example of the use of the GR scheme
here (figure~\ref{ex-sent}) rather than duplicating Carroll, Briscoe \&
Sanfilippo's description of it. 
\begin{figure}
When the proprietor dies, the establishment should become a corporation
until it is either acquired by another proprietor or the government
decides to drop it.
\begin{quote}
cmod(when, become, die)\\
ncsubj(die, proprietor, \_)\\
ncsubj(become, establishment, \_)\\
xcomp(become, corporation, \_)\\
mod(until, become, acquire)\\
ncsubj(acquire, it, obj)\\
arg\_mod(by, acquire, proprietor, subj)\\
cmod(until, become, decide)\\
ncsubj(decide, government, \_)\\
xcomp(to, decide, drop)\\
ncsubj(drop, government, \_)\\
dobj(drop, it, \_)
\end{quote}

\caption{Example sentence and GRs (\susanne\ rel3, lines
G22:1460k--G22:1480m).}
\label{ex-sent}
\end{figure}
The set of possible relations (i.e.\ {\it cmod}, {\it ncsubj}, etc.)\ is
organised hierarchically; see figure~\ref{grs-used}.
The most generic relation between a head and a dependent is {\it dependent}.
Where the relationship between the two is known more precisely, relations
further down the hierarchy can be used, for example {\it mod}(ifier) or {\it
arg}(ument). Relations {\it mod}, {\it arg\_mod}, {\it clausal}, and their
descendants have slots filled by a {\small\bf type}, a {\small\bf head}, and its {\small\bf
dependent}; {\it arg\_mod} has an additional fourth slot {\small\bf initial\_gr}.
Descendants of {\it subj}, and also {\it dobj} have the three slots {\small\bf
head}, {\small\bf dependent}, and {\small\bf initial\_gr}. The {\it x} and
{\it c} prefixes to relation names differentiate clausal control alternatives.

The scheme is
superficially similar to a syntactic dependency analysis in the style of Lin
(1995). However, the scheme contains a specific, fixed inventory of relations.
Other significant differences are:
\begin{itemize}
\item the GR analysis of control relations could not be expressed as a strict
dependency tree since a single nominal head would be a dependent of two (or
more) verbal heads (as with {\it ncsubj(decide, government, \_) ncsubj(drop,
government, \_)} in the figure~\ref{ex-sent} example {\it ...the government
decides to drop it});

\item any complementiser or preposition linking a head with a clausal or
PP dependent is an integral part of the GR (the {\small\bf type} slot); 

\item the underlying grammatical relation is specified for arguments
``displaced'' from their canonical positions by movement phenomena
(e.g.\ the {\small\bf initial\_gr} slot of {\it ncsubj} and {\it arg\_mod}
in the passive {\it ...it is either acquired by another proprietor...});

\item semantic arguments syntactically realised as modifiers (e.g.\
the passive by-phrase) are indicated as such---using {\it arg\_mod};

\item conjuncts in a co-ordination structure are distributed over the
higher-level relation (e.g.\ in {\it ...become ... until ... either acquired
... or ... decides...} there are two verbal dependents of {\it
become}, {\it acquire} and {\it decide}, each in a separate {\it mod} GR;

\item arguments which are not lexically realised can be expressed (e.g.\ when
there is pro-drop the dependent in a {\it subj} GR would be specified as {\small\bf
Pro});

\item GRs are organised into a hierarchy so that they can be left
underspecified by a shallow parser which has incomplete knowledge
of syntax.
\end{itemize}
In addition to constituent structure, both the \ptb\ and \susanne\ contain
functional, or pred\-icate-argument annotation, the former particularly
employing a rich set of distinctions, often with complex grammatical and
contextual conditions on when one function tag should be applied in preference
to another. For example, the tag {\small TPC} (``topicalized'')
\begin{quote}
``--- marks elements that appear before the subject in a declarative sentence,
but in two cases only: (i) if the fronted element is associated with a *T* in
the position of the gap. (ii) if the fronted element is left-dislocated [...]''
\end{quote}
(Bies \etal, 1995: 40). Conditions of this type would be very difficult to
encode in an actual parser, so attempting to evaluate on them would be
uninformative. Much of the problem is that treebanks of this kind have to
specify the behaviour of many interacting factors, such as how syntactic
constituents should be segmented, labelled and structured hierarchically, how
displaced elements should be co-indexed, and so on. Within such a framework
the further specification of how functional tags should be attached to
constituents is necessarily highly complex. Moreover, functional information is
in some cases left implicit\footnote{``The predicate is the lowest (right-most
branching) VP or (after copula verbs and in `small clauses') a constituent
tagged PRD'' (Bies \etal, 1995: 11).}, presenting further problems for precise
evaluation. Table~\ref{funct-comparison} gives a rough comparison between the
types of information in the GR scheme and in the \ptb\ and \susanne. It might
be possible semi-automatically to map a treebank predicate-argument
encoding to the GR scheme (taking advantage of the large amount of work that
has gone into the treebanks), but we have not investigated this to date.
\begin{table}[tb]
\centering
\begin{tabular}{|l|cc|} \hline
   \multicolumn{1}{|l|}{Relation}
   & \multicolumn{1}{c}{\ptb}
   & \multicolumn{1}{c|}{\susanne} \\[1mm]
\hline

{\it dependent}        &  -- &  --\\
~{\it mod}             &  {\small TPC/ADV} etc.\ &  p etc.\\
~~{\it ncmod}          &  {\small CLR/VOC/ADV} etc.\ & n/p etc.\\
~~{\it xmod}           &  &   \\
~~{\it cmod}           &  &   \\
~{\it arg\_mod}        &  {\small LGS} &   a\\
~{\it arg}             &  -- & --\\
~~{\it subj}           &  -- &  --\\
~~~{\it ncsubj}        &  {\small SBJ} &  s\\
~~~{\it xsubj}         &  &  \\
~~~{\it csubj}         &  &  \\
~~{\it subj\_or\_dobj} &  -- &  --\\
~~{\it comp}           &  -- &  --\\
~~~{\it obj}           &  -- &  --\\
~~~~{\it dobj}         &  ({\small NP} after {\small V}) &   o\\
~~~~{\it obj2}         &  (2nd {\small NP} after {\small V}) &   \\
~~~~{\it iobj}         &  {\small CLR/DTV} &   i\\
~~~{\it clausal}       &  {\small PRD} &  --\\
~~~~{\it xcomp}        &  &   e\\
~~~~{\it ccomp}        &  &   j\\ \hline
\end{tabular}

\caption{Rough correspondence between the GR scheme and the functional
annotation in the Penn Treebank (\ptb) and \susanne.}
\label{funct-comparison}
\end{table}

\section{The Annotated Corpus and Evaluation}

\subsection{Corpus Annotation}

Our corpus consists of 500 sentences (10K words) 
covering a number of written genres. The sentences were taken from the
\susanne\ corpus, and each was marked up manually by two
annotators\footnote{The corpus and evaluation software that can be used with
it will shortly be made publicly available online.}. 

The manual analysis was performed by the first author and was checked
and extended by the third author. Inter-annotator agreement was around
95\% which is somewhat better than previously reported figures
for syntactic markup
(e.g.\ Leech and Garside, 1991). Marking up was done
semi-automatically by first generating the set of relations predicted
by the evaluation software from the closest system analysis to the
treebank annotation and then manually correcting and extending
these.

The mean number of GRs
per corpus sentence is 9.72. Table~\ref{grs-comparison} quantifies the
distribution of relations occurring in the corpus. 
\begin{table}
\centering
\begin{tabular}{|l|rr|} \hline
   \multicolumn{1}{|l|}{Relation}
   & \multicolumn{1}{c}{\# occurrences}
   & \multicolumn{1}{c|}{\% occurrences} \\[1mm]
\hline

{\it dependent}        & 4690 & 100.0\\
~{\it mod}             & 2710 &  57.8\\
~~{\it ncmod}          & 2377 &  50.7\\
~~{\it xmod}           &  170 &   3.6\\
~~{\it cmod}           &  163 &   3.5\\
~{\it arg\_mod}        &   39 &   0.8\\
~{\it arg}             & 1941 &  41.4\\
~~{\it subj}           &  993 &  21.2\\
~~~{\it ncsubj}        &  984 &  21.0\\
~~~{\it xsubj}         &    5 &   0.1\\
~~~{\it csubj}         &    4 &   0.1\\
~~{\it subj\_or\_dob}j & 1339 &  28.6\\
~~{\it comp}           &  948 &  20.2\\
~~~{\it obj}           &  559 &  11.9\\
~~~~{\it dobj}         &  396 &   8.4\\
~~~~{\it obj2}         &   19 &   0.4\\
~~~~{\it iobj}         &  144 &   3.1\\
~~~{\it clausal}       &  389 &   8.3\\
~~~~{\it xcomp}        &  323 &   6.9\\
~~~~{\it ccomp}        &   66 &   1.4\\ \hline
\end{tabular}

\caption{Frequency of each type of GR (inclusive of subsumed relations) in
the 10K-word corpus.}
\label{grs-comparison}
\end{table}
The split between modifiers
and arguments is roughly 60/40, with approximately equal numbers of subjects
and complements. Of the latter, 40\% are clausal; clausal modifiers are
almost as prevalent. In strong contrast, clausal subjects are highly
infrequent (accounting for only 0.2\% of the total). Direct objects
are 2.75 times more frequent than indirect objects, which are themselves
7.5 times more prevalent than second objects.

The corpus contains sentences belonging to three distinct genres.
These are classified in the original Brown corpus as: {\it A}, press
reportage; {\it G}, belles lettres; and {\it J}, learned writing.
Genre has been found to affect the distribution of surface-level
syntactic configurations (Sekine, 1997) and also complement types
for individual predicates (Roland \& Jurafsky, 1998). However, we observe no
statistically significant difference in the {\it total} numbers of the
various grammatical relations across the three genres in the corpus.

\subsection{Parser Evaluation}

We replicated an experiment previously
reported by Carroll, Minnen \& Briscoe (1998), using a robust lexicalised
parser, computing three evaluation measures for each type of relation against
the 10K-word test corpus (table~\ref{subcat-relation}).
The evaluation measures are precision, recall, and
F-score (van Rijsbergen, 1979)\footnote{The F-score is a measure combining
precision and recall into a single figure. We use the version in which they
are weighted equally, defined as $2 \times precision \times recall/(precision
+ recall)$.} of parser GRs against the test corpus annotation.
\begin{table}
\centering
\begin{tabular}{|l|rrr|} \hline
Relation & \multicolumn{1}{c}{Precision} & \multicolumn{1}{c}{Recall}
         & \multicolumn{1}{c|}{F-score} \\
         & \multicolumn{1}{c}{(\%)} & \multicolumn{1}{c}{(\%)}
         & \multicolumn{1}{c|}{} \\
\hline

{\it dependent}        & 75.1           & 75.2            & 75.1\\
~{\it mod}             & 73.7           & 69.7            & 71.7\\
~~{\it ncmod}          & 78.1           & 73.1            & 75.6\\
~~{\it xmod}           & 70.0           & 51.9            & 59.6\\
~~{\it cmod}           & 67.4           & 48.1            & 56.1\\
~{\it arg\_mod}        & 84.2           & 41.0            & 55.2\\
~{\it arg}             & 76.6           & 83.5            & 79.9\\
~~{\it subj}           & 83.6           & 87.9            & 85.7\\
~~~{\it ncsubj}        & 84.8           & 88.3            & 86.5\\
~~~{\it xsubj}        & 100.0           & 40.0            & 57.1\\
~~~{\it csubj}         & 14.3          & 100.0            & 25.0\\
~~{\it subj\_or\_dobj} & 84.4           & 86.9            & 85.6\\
~~{\it comp}           & 69.8           & 78.9            & 74.1\\
~~~{\it obj}           & 67.7           & 79.3            & 73.0\\
~~~~{\it dobj}         & 86.3           & 84.3            & 85.3\\
~~~~{\it obj2}         & 39.0           & 84.2            & 53.3\\
~~~~{\it iobj}         & 41.7           & 64.6            & 50.7\\
~~~{\it clausal}       & 73.0           & 78.4            & 75.6\\
~~~~{\it xcomp}        & 84.4           & 78.9            & 81.5\\
~~~~{\it ccomp}        & 72.3           & 74.6            & 73.4\\ \hline
\end{tabular}
\caption{GR accuracy by relation.}
\label{subcat-relation}
\end{table}

GRs are in general compared using an
equality test, except that we allowed the parser to return {\it
mod}, {\it subj} and {\it clausal} relations rather than the more specific
ones they subsume, and to leave unspecified the filler for the {\small\bf
type} slot in the {\it mod}, {\it iobj} and {\it clausal}
relations\footnote{The implementation of the
extraction of GRs from parse trees is currently being refined, so
these minor relaxations should be removed soon.}. The {\small\bf head} and
{\small\bf dependent} slot fillers are in all cases the base forms of single
head words, so for example, `multi-component' heads such as the names of
people and companies are reduced to a single word; thus the slot filler
corresponding to {\it Bill Clinton} would be {\it Clinton}. For real-world
applications this might not be the desired behaviour---one might instead
want the token {\it Bill\_Clinton}---but the analyser could easily be
modified to do this.

The evaluation results can be used to give a single figure for parser
accuracy---the F-score of the {\it dependent} relation---precision and
recall at the most general level, or more fine-grained information about how
accurately groups of, or single relations were produced. The latter would be
particularly useful during parser/ grammar development to identify where
effort should be expended on making improvements.

\section{Conclusions}

We have outlined and justified a language and applic\-ation-independent corpus
annotation scheme for evaluating syntactic parsers, based on grammatical
relations between heads and dependents. The scheme has been used in the
EU-funded SPARKLE project (see $<$http://www.ilc.pi.cnr.it/sparkle.html$>$) to
annotate English, French, German and Italian corpora, and for evaluating
parsers for these languages. In this paper we have described a 10K-word corpus
of English marked up to this standard, and shown its use in
evaluating a robust parsing system. The corpus and evaluation software that
can be used with it will shortly be made publicly available online.

\section*{Acknowledgments}

This work was funded by UK EPSRC project GR/L53175
`PSET: Practical Simplification of English Text', CEC Telematics
Applications Programme project LE1-2111 `SPARKLE: Shallow PARsing and
Knowledge extraction for Language Engineering', and by an EPSRC Advanced
Fellowship to the first author. We would like to thank Antonio
Sanfilippo for his substantial input to the design of the annotation scheme.

\section*{References}

\newcommand{\book}[4]{\item #1 (#4) {\it #2}. #3.}
\newcommand{\barticle}[7]{\item #1 (#7) #2. In #5 (Eds.), {\it #4},
#3. #6.}
\newcommand{\bparticle}[6]{\item #1 (#6) #2. In #4 (Eds.), {\it #3}. #5.}
\newcommand{\boarticle}[5]{\item #1 (#5) #2. In {\it #3}. #4.}
\newcommand{\farticle}[6]{\item #1 (#6) #2. In #4 (Eds.), {\it #3}: #5.
Forthcoming.}
\newcommand{\uarticle}[5]{\item #1 (#5) #2. In #4 (Eds.), {\it #3}.
Forthcoming.}
\newcommand{\jarticle}[6]{\item #1 (#6) #2. {\it #3}, #4, #5.}
\newcommand{\particle}[6]{\item #1 (#6) #2. In {\it Proceedings of the
#3}, #5. #4.}
\newcommand{\lazyparticle}[5]{\item #1 (#5) #2. In {\it Proceedings of the
#3}. #4.}
\newcommand{\lazyjarticle}[4]{\item #1 (#4) #2. {\it #3}.}
\newcommand{\lazyfjarticle}[4]{\item #1 (#4) #2. {\it #3}. Forthcoming.}
\newcommand{\bookartnopp}[6]{\item #1 (#6) #2. In #4 (Eds.), {\it #3,} #5.}

\begin{list}{}
   {\leftmargin 15pt
    \itemindent -15pt
    \itemsep 0pt plus 0pt
    \parsep 1pt plus 0pt}

\barticle{Atwell, E.}
{Comparative evaluation of grammatical annotation models}
{25--46}
{Industrial Parsing of Software Manuals}
{R. Sutcliffe, H. Koch \& A. McElligott}
{Amsterdam: Rodopi}
{1996}

\book{Bies, A., Ferguson, M., Katz, K., MacIntyre, R., Tredinnick, V., Kim,
G., Marc\-in\-kie\-wicz, M., Schasberger, B.}
{Bracketing guidelines for Treebank II style Penn Treebank Project}
{Technical Report, CIS, University of Pennsylvania, Philadelphia, PA}
{1995}

\lazyparticle{Carpenter, B. \& Manning, C.}
{Probabilistic parsing using left corner language models}
{5th ACL/SIGPARSE International Workshop on Parsing Technologies}
{MIT, Cambridge, MA}
{1997}

\particle{Carroll, J., Briscoe E. \& Sanfilippo, A.}
{Parser evaluation: a survey and a new proposal}
{International Conference on Language Resources and Evaluation}
{Granada, Spain. Available online at
$<$ftp://ftp.cogs.susx.ac.uk/pub/users/johnca/ lre98-final.ps$>$}
{447--454}
{1998}

\lazyparticle{Carroll, J., Minnen, G. \& Briscoe E.}
{Can subcategorisation probabilities help a statistical parser?}
{6th ACL/SIGDAT Workshop on Very Large Corpora}
{Montreal, Canada. Available online at
$<$http://xxx.lanl.gov/abs/cmp-lg/9806013$>$}
{1998}

\particle{Charniak, E.}
{Tree-bank grammars}
{13th National Conference on Artificial Intelligence, AAAI-96}
{}
{1031--1036}
{1996}

\particle{Collins, M.}
{A new statistical parser based on bigram lexical dependencies}
{34th Meeting of the Association for Computational Linguistics}
{Santa Cruz, CA}
{184--191}
{1996}

\lazyjarticle{Gaizauskas, R.}
{Special issue on evaluation}
{Computer Speech \& Language}
{1998}

\lazyparticle{Gaizauskas, R., Hepple M. \& Huyck, C.}
{Modifying existing annotated corpora for general comparative
evaluation of parsing}
{LRE Workshop on Evaluation of Parsing Systems}
{Granada, Spain}
{1998}

\particle{Grishman, R., Macleod, C. \& Sterling, J.}
{Evaluating parsing strategies using standardized parse files}
{3rd ACL Conference on Applied Natural Language Processing}
{Trento, Italy}
{156--161}
{1992}
 
\lazyparticle{Harrison, P., Abney, S., Black, E., Flickinger, D.,
Gdaniec, C., Grishman, R., Hindle, D., Ingria, B., Marcus, M.,
Santorini, B. \& Strzalkowski, T.}
{Evaluating syntax performance of parser/grammars of English}
{Workshop on Evaluating Natural Language Processing Systems}
{ACL}
{1991}

\bookartnopp{Leech, G. \& Garside, R.}
{Running a grammar factory: the production of syntactically analysed corpora
or `treebanks'}
{English Computer Corpora: Selected Papers and Bibliography}
{S. Johansson \& A. Stenstrom}
{Mouton de Gruyter, Berlin}
{1991}

\particle{Lehmann, S., Oepen, S., Regnier-Prost, S., Netter, K.,
Lux, V., Klein, J., Falkedal, K., Fouvry, F., Estival, D., Dauphin, E.,
Compagnion, H., Baur, J., Balkan, L. \& Arnold, D.}
{{\sc tsnlp} --- test suites for natural language processing}
{International Conference on Computational
Linguistics, COLING-96}
{Copenhagen, Denmark}
{711--716}
{1996}

\particle{Lin, D.}
{A dependency-based method for evaluating broad-coverage parsers}
{14th International Joint Conference on Artificial Intelligence}
{Montreal, Canada}
{1420--1425}
{1995}

\barticle{Lin, D.}
{Dependency-based parser evaluation: a study with a software manual corpus}
{13--24}
{Industrial Parsing of Software Manuals}
{R. Sutcliffe, H-D. Koch \& A. McElligott}
{Amsterdam, The Netherlands: Rodopi}
{1996}

\jarticle{Marcus, M., Santorini, B. \& Marcinkiewicz}
{Building a large annotated corpus of English: The Penn Treebank}
{Computational Linguistics}
{19(2)}
{313--330}
{1993} 

\particle{Roland, D. \& Jurafsky, D.}
{How verb subcategorization frequencies are affected by corpus choice}
{17th International Conference on Computational Linguistics (COLING-ACL'98)}
{Montreal, Canada}
{1122--1128}
{1998} 

\book{Rubio, A. (Ed.)}
{International Conference on Language Resources and Evaluation}
{Granada, Spain}
{1998}

\book{Sampson, G.} {English for the Computer} {Oxford, UK: Oxford University
Press} {1995}

\book{Sanfilippo, A., Barnett, R., Calzolari, N., Flores, S., Hellwig, P.,
Leech, P., Melero, M., Montemagni, S., Odijk, J., Pirrelli, V.,
Teufel, S., Villegas M. \& Zaysser, L.}
{Subcategorization Standards}
{Report of the EAGLES Lexicon/Syntax Interest Group. Available through
{\tt eagles@ilc.pi.cnr.it}}
{1996}

\particle{Sekine, S.}
{The domain dependence of parsing}
{5th ACL Conference on Applied Natural Language Processing}
{Washington, DC}
{96--102}
{1997}

\lazyparticle{Srinivas, B., Doran, C., Hockey B. \& Joshi A.}
{An approach to robust partial parsing and evaluation metrics}
{ESSLLI'96 Workshop on Robust Parsing}
{Prague, Czech Republic}
{1996}

\lazyparticle{Srinivas, B., Doran, C. \& Kulick, S.}
{Heuristics and parse ranking}
{4th ACL/SIGPARSE International Workshop on Parsing Technologies}
{Prague, Czech Republic}
{1995}

\book{van Rijsbergen, C.} {Information Retrieval} {Butterworth, London} {1979}

\end{list}

\end{document}